\documentclass[preprint,11pt]{imsart}

\usepackage{amsthm,amsmath,amssymb}
\usepackage[numbers]{natbib}
\usepackage[colorlinks,citecolor=blue,urlcolor=blue]{hyperref}

\usepackage[margin=2.5cm]{geometry}

\startlocaldefs
\numberwithin{equation}{section}
\theoremstyle{plain}
\newtheorem{theorem}{Theorem}
\newtheorem{assume}{Assumption}
\newtheorem{lemma}{Lemma}
\newtheorem{cor}{Corollary}
\newtheorem{prop}{Proposition}
\newtheorem{rmk}{Remark}
\DeclareMathOperator*{\argmin}{arg\,min}

\endlocaldefs

\allowdisplaybreaks

\begin{document}
	\begin{frontmatter}
		\title{A sparse PAC-Bayesian approach for high-dimensional quantile prediction}
	\runtitle{PAC-Bayesian high-dimensional quantile prediction}
		
\begin{aug}
	\author{\fnms{The Tien}~\snm{Mai}\ead[label=e1]{the.t.mai@ntnu.no}\orcid{0000-0002-3514-9636}}	
\address{
	Department of Mathematical Sciences,
\\
Norwegian University of Science and Technology,
	Trondheim 7034, Norway.
\\
	\printead[presep={\ }]{e1}
}
\runauthor{Mai T. T.}
\end{aug}

\begin{abstract}	
Quantile regression, a robust method for estimating conditional quantiles, has advanced significantly in fields such as econometrics, statistics, and machine learning. In high-dimensional settings, where the number of covariates exceeds sample size, penalized methods like lasso have been developed to address sparsity challenges. Bayesian methods, initially connected to quantile regression via the asymmetric Laplace likelihood, have also evolved, though issues with posterior variance have led to new approaches, including pseudo/score likelihoods. This paper presents a novel probabilistic machine learning approach for high-dimensional quantile prediction. It uses a pseudo-Bayesian framework with a scaled Student-t prior and Langevin Monte Carlo for efficient computation. The method demonstrates strong theoretical guarantees, through PAC-Bayes bounds, that establish non-asymptotic oracle inequalities, showing minimax-optimal prediction error and adaptability to unknown sparsity. Its effectiveness is validated through simulations and real-world data, where it performs competitively against established frequentist and Bayesian techniques.
\end{abstract}

		

\end{frontmatter}

\section{Introduction}

In a wide range of research domains, such as genomics, economics, and finance, high-dimensional data are frequently collected. Analyzing these data sets introduces significant challenges for statisticians, prompting the need for novel statistical methods and theories \citep{buhlmann_vandegeer,wainwright2019high,giraud2021introduction}. In this work, we address the high-dimensional quantile prediction problem, specifically when the number of covariates is greater than the sample size.

Quantile regression has played a significant role in various disciplines, including econometrics, statistics, and machine learning, ever since the foundational work of \cite{koenker1978regression}. This importance has evidenced by the ongoing developments and comprehensive reviews provided by \cite{koenker2017quantile}. Quantile regression is recognized as a robust statistical method that estimates a conditional quantile of interest, making it particularly useful for understanding the distributional effects of covariates on different points of the outcome variable, rather than focusing solely on the mean. In the context of high-dimensional data analysis, where the number of covariates often exceeds the sample size, sparsity structures have been identified as a crucial condition for effectively analyzing quantile regression models.  To address these challenges, various penalized quantile regression approaches have been developed. Among these, the use of the Lasso penalty, as explored in \cite{belloni2011ell_1}, has been influential in promoting sparsity by shrinking some coefficients exactly to zero. Additionally, the $\ell_0$ approach, which focuses on selecting a subset of covariates, has been considered in \cite{chen2023sparse}, providing an alternative method for handling high-dimensional quantile regression.

Moreover, nonconvex penalties, as discussed in \cite{wang2012quantile}, have been employed to estimate linear quantile regression models with high-dimensional covariates, offering a more flexible approach to model complex settings. The high-dimensional linear quantile regression model has also been extended to more flexible frameworks, such as partially linear models \cite{sherwood2016variable,sherwood2016partially} and nonparametric models \cite{sherwood2022additive}. These extensions allow for greater model adaptability by incorporating non-linear relationships and interactions among covariates, further enhancing the applicability of quantile regression in high-dimensional scenarios.

The Bayesian approach to quantile regression was first proposed by \cite{yu2001bayesian}, relying on the asymmetric Laplace likelihood (ALL). This likelihood function is particularly noteworthy due to its strong connection to the frequentist quantile regression framework; specifically, the maximum likelihood estimates derived from the ALL are equivalent to the traditional quantile regression estimates obtained using the check-loss function \cite{koenker2005quantile}. Despite this appealing connection, subsequent research by \cite{sriram2015sandwich,yang2016posterior} identified a critical issue: the asymptotic variance of the posterior distribution based on the ALL is incorrect. This discrepancy implies that the inferences drawn from the Bayesian quantile regression approach may be invalid when viewed from a frequentist perspective. To address this problem, these authors proposed a correction to the posterior variance, thereby enabling valid inference under the Bayesian quantile regression framework. Following this development, a variety of alternative pseudo Bayesian approaches have been explored for quantile regression: \cite{yang2012bayesian,xi2016bayesian} studied a Bayesian empirical likelihood method, \cite{li2010bayesian} proposed a regularized Bayesian approach, \cite{Li2024PSEUDO} studied a pseudo Bayesian framework. Moreover, in the work of \cite{wu2021bayesian}, the authors introduced the concept of score likelihood, which provides a further alternative to the ALL-based approach. This diversity of methodologies reflects ongoing efforts within the Bayesian community to enhance the robustness and reliability of quantile regression, addressing the limitations identified in the original ALL-based framework.

Current theoretical frameworks for Bayesian methods are primarily focused on asymptotic results, which rely on the assumption of large sample sizes and often fail to address high-dimensional settings. In these high-dimensional scenarios, conventional asymptotic assumptions no longer hold, creating a gap in the literature regarding the application of Bayesian methods under such challenging conditions. To our knowledge, there are no well-established theoretical results that specifically address the use of Bayesian methods in high-dimensional contexts. Additionally, there is a lack of results on the predictive performance of Bayesian methods in quantile regression.

Recognizing this significant gap, our paper aims to contribute to the field by introducing a novel methodological approach designed to address these high-dimensional challenges. This approach is informed by and builds upon the insights from previous works, including \cite{xi2016bayesian,wu2021bayesian,Li2024PSEUDO}, which have advocated for innovative strategies to overcome the limitations of traditional Bayesian methods. Specifically, these prior studies propose substituting the conventional likelihood-based approach with a loss/score or risk function. This alternative framework shifts the focus from the likelihood function to a more flexible and robust criterion that can better handle the complexity of the data. By adopting this innovative perspective, our paper seeks to extend the theoretical underpinnings of Bayesian methods to high-dimensional settings, offering a more comprehensive framework that is both theoretically sound and practically applicable.

Our approach diverges from traditional parametric models by focusing on a set of predictors, with the goal of identifying the one that minimizes prediction error. We employ a probabilistic machine learning methodology by incorporating a Gibbs posterior distribution, introducing a pseudo-Bayesian method tailored for high-dimensional quantile prediction. This approach utilizes a risk notion based on the quantile loss, rather than a likelihood function, and is grounded in the principles of PAC-Bayesian theory \citep{STW,McA,herbrich2002pac,catonibook,dalalyan2008aggregation,seldin2010pac,germain2015risk,alquier2016}. Additionally, the use of loss functions in place of likelihoods, as in \cite{xi2016bayesian,wu2021bayesian,Li2024PSEUDO}, has gained popularity in generalized Bayesian inference, as evidenced by recent literature \citep{matsubara2022robust,jewson2022general,yonekura2023adaptation,medina2022robustness,grunwald2017inconsistency,bissiri2013general,lyddon2019general,syring2019calibrating}.

We provide non-asymptotic excess risk bounds for our proposed approach, demonstrating that the prediction errors of our proposed methods reach the minimax-optimal rate and comparable to those established in the frequentist literature, such as in \cite{belloni2011ell_1,wang2019l_1,chen2023sparse}. A fundamental strategy for addressing challenges in high-dimensional settings is to impose sparsity within the prior distribution, and in our approach, we specifically utilize a scaled Student-t prior distribution, introduced by \cite{dalalyan2012mirror}. This choice is particularly advantageous when dealing with high-dimensional data, where sparsity is essential for effective modeling and inference. While other continuous shrinkage priors have been considered for sparse quantile regression, such as the Horseshoe prior \cite{kohns2024horseshoe}, it is important to note that the theoretical validation of these priors has not yet been established for quantile regression. In contrast, the employment of the scaled Student-t prior in our framework offers a more theoretically grounded approach. 

Additionally, the use of the scaled Student-t prior enables us to leverage the Langevin Monte Carlo (LMC) method, a sampling technique that effectively utilizes gradient information. The LMC method has shown considerable promise in high-dimensional Bayesian applications, as evidenced by the theoretical and empirical studies in \cite{dalalyan2017theoretical,durmus2017nonasymptotic,durmus2019high}. We have also thoroughly evaluated the effectiveness and robustness of our proposed method through comprehensive numerical studies.

First, we applied our method to simulated data to evaluate its performance under controlled conditions. We then conducted a detailed empirical analysis on two distinct applications: predicting the fat content of meat based on a 100-channel spectrum of absorbance measurements, and predicting the expression levels of the TRIM32 gene using gene expression data from rat eyes.

To rigorously validate our method, we performed a comparative analysis against two state-of-the-art methodologies. The first comparison was with a frequentist approach using Lasso penalization, which is well-established for handling high-dimensional data by imposing sparsity through regularization. The second comparison involved a recently introduced Bayesian sparse method for quantile regression, which utilizes the Horseshoe prior, as detailed by \cite{kohns2024horseshoe}. Through both simulation studies and real data analyses, we demonstrated that our proposed method performs at least on par with these advanced techniques, offering a competitive, reliable, and versatile alternative across various types of data.

The remainder of the paper is organized as follows. Section \ref{sc_modelmethod} presents the problem statement and introduces our approach using a Gibbs posterior. Section \ref{sc_theory} details our main theoretical findings regarding prediction errors. Section \ref{sc_numerical} covers the simulation studies, while Section \ref{sc_realdata} discusses applications to real data. The paper concludes in Section \ref{sc_conclusion}, and all technical proofs are provided in Appendix \ref{sc_proofs}.

\section{Model and method}
\label{sc_modelmethod}
{\bf Notations:}
Let $P $ and $ R $ be two probability distributions and let $\mu$ be any measure such that $P\ll \mu$ and $R\ll \mu$.   The Kullback-Leibler divergence is defined by
$
\mathcal{K}(P,R)  = 
\int \log \left(\frac{{\rm d}P}{{\rm d}R} \right){\rm d}P $  if  $ P \ll R
$, and  $
+ \infty $  otherwise. Let $ \|\cdot\|_q $ denote the $ \ell_q $-norm, $ \|\cdot\|_\infty $ denote the max-norm of vectors, and let $ \|\cdot\|_0 $ denote the $ \ell_0 $ (quasi)-norm (the number nonzero entries) of vectors.

\subsection{Problem statement}

Let $ Z_i:= (Y_i,X_i) \in \mathbb{R} \times \mathbb{R}^d $, $ i=1,\ldots,n $, be a collection of $n$ independent and identically distributed (i.i.d) random samples drawn from a distribution $ \mathbf{P} $. Suppose $ \theta \in \mathbb{R}^d $ is a linear predictor. We focus on  the $ \tau $-th quantile prediction, where $ \tau \in (0,1) $: that given $ (Y, X) \sim \mathbf{P} $, the prediction for $ Y $ is made by $ X^\top \theta $. The ability of this predictor to predict a new $ Y $ is then assessed by the quantile risk
\begin{align*}
R (\theta) = \mathbb{E}_{\mathbf{P}}
[ \ell_\tau (Y, X^\top \theta) ]
\end{align*}
where $ \ell_\tau (y,x) = (y-x) (\tau - 1_{\{ y \leq x \} } ) $, and the empirical quantile risk is given as
\begin{align*}
r_n (\theta) 
= 
\frac{1}{n} \sum_{i=1}^{n}
 \ell_\tau (Y_i , X_i^\top \theta) 
 .
\end{align*}

Define 
$
\theta^* 
= \argmin_{ \theta\in\mathbb{R}^d } R (\theta) 
.
$
We consider in this paper the prediction problem, i.e., build an estimator $ \hat{\theta} $ from the data $ (Y_i,X_i)_{i=1,\ldots, n } $ such that $  R(\hat{\theta}) $ is close to $ R^* := \min_{ \theta\in\mathbb{R}^d } R (\theta)  $ up to a positive remainder term as small as possible.

A high-dimensional sparse scenario is studied in this paper: we assume that $ s^* < n < d $, where $ s^*:= \|\theta^* \|_0 $.

\subsection{A sparse Gibbs posterior approach}

The choice of prior distribution plays a critical role in ensuring an accurate prediction error rate in high-dimensional models. 
For a fixed constant \( C_1 >0 \), put \( B_1(C_1) := \{ \theta \in \mathbb{R}^d : \|\theta\|_1 \leq C_1 \} \). For each \( \theta \) within \( B_1(C_1) \), we adopt the scaled Student distribution as our prior distribution.
\begin{eqnarray}
\label{eq_priordsitrbution}
\pi (\theta) 
\propto 
\prod_{i=1}^{d} 
(\varsigma^2 + \theta_{i}^2)^{-2}
,
\end{eqnarray}
where $ \varsigma>0 $ is a tuning parameter.  This prior has been applied in various sparse settings, as demonstrated in \cite{dalalyan2012mirror, dalalyan2012sparse, mai2023high}. In this context, $C_1$ acts as a regularization parameter and is generally assumed to be very large. As a result, the distribution of $\pi $ approximates that of $S\varsigma \sqrt{2}$, where $S$ is a random vector with i.i.d. components drawn from a Student's t-distribution with 3 degrees of freedom. By choosing a very small value for $\varsigma$, the majority of elements in $\varsigma S$ are concentrated near zero. However, because of the heavy-tailed nature, a small proportion of components in $\varsigma S$ significantly deviate from zero. This property allows the prior to effectively encourage sparsity in the parameter vector. The significance of heavy-tailed priors in fostering sparsity has been explored in prior studies as in \citep{seeger2008bayesian, johnstone2004needles, rivoirard2006nonlinear, abramovich2007optimality, carvalho2010horseshoe, castillo2012needles, castillo2015bayesian, castillo2018empirical, ray2022variational}.

For any \(\lambda > 0\), as in the PAC-Bayesian framework \cite{catonibook,alquier2021user}, the Gibbs posterior \(\hat{\rho}_{\lambda} \) is defined by
	\begin{equation}
\label{eq_Gibbs_poste}
\hat{\rho}_{\lambda} (\theta) 
= 
\frac{\exp[-\lambda r_n(\theta)]}
{\int \exp[-\lambda r_n ] {\rm d}\pi } \pi(\theta)
, 	
\end{equation}
and let $ \hat{\theta} = \int \theta \hat{\rho}_{\lambda} ({\rm d}\theta)  $ be our mean estimator.

The Gibbs posterior in \eqref{eq_Gibbs_poste} is also known as the EWA (exponentially weighted aggregate) procedure, as referenced in \cite{alquier2016,catonibook,dalalyan2012sparse,dalalyan2008aggregation}. The use of $\hat{\rho}_\lambda $ is driven by the minimization problem outlined in Lemma \ref{lemma:dv}, rather than strictly following traditional Bayesian methods. Importantly, there is no requirement for a likelihood function or a complete model; the focus is solely on the empirical risk derived from the quantile loss function. In this paper, we consistently use $\pi$ to represent the prior and $\hat{\rho}_\lambda$ to denote the pseudo-posterior. The purpose of the EWA is to adjust the distribution to favor parameter values with lower in-sample quantile empirical risk, with the degree of adjustment governed by the tuning parameter $\lambda$, which will be explored further in the following sections.

\section{Theoretical guarantee}
\label{sc_theory}
\subsection{Assumption}

We present the assumptions that are essential for obtaining our theoretical findings.

\begin{assume}
	\label{assume_X_bounded}
	We assume that there exists a constant $ C_{\rm x} >0 $ such that
	$ 	\mathbb{E} \|X_1\| \leq C_{\rm x} < \infty $.
\end{assume}

Our results primarily rely on Assumption \ref{assume_X_bounded}, which was also utilized by \cite{jiang2007bayesian} in the context of regression problems involving random design.

\begin{assume}
	\label{assum_bernstein}
	Assume that for any $\theta\in\Theta $, $ R(\theta) \geq R(\theta^*) $  and there is a constant $K>0 $ such that, 
	\begin{equation*}
	\mathbb{E} \{ 
	|  X^\top (\theta - \theta^*) |^2 \}
	\leq 
	K [ R (\theta) - R(\theta^*) ]
	,
	\end{equation*}
\end{assume}

Assumption \ref{assum_bernstein}, which is central to \cite{chen2023sparse} for sparse quantile regression, can be regarded as a type of Bernstein's condition. This condition has been extensively studied in the learning theory literature, for example in \citep{mendelson2008obtaining,zhang2004statistical,alquier2019estimation,elsener2018robust}. For our purposes, this assumption is just important to achieving a fast rate.

\begin{assume}
	\label{assum_eigen}
	Assume that there is a constant $ \kappa >0 $ such that, for any $\theta $ that $ \|\theta \|_0 \leq s^* $,
	\begin{equation*}
	\kappa^2	\|  \theta - \theta^* \|^2
	\leq
	\mathbb{E} \{ 
	|  X^\top (\theta - \theta^*) |^2 \}
	,
	\end{equation*}
\end{assume}

Assumption \ref{assum_eigen} has also been utilized in \cite{chen2023sparse}. This assumption is satisfied when the smallest eigenvalue of \( \mathbb{E} (XX^\top) \) is bounded below by a positive constant, which is similar to the sparse eigenvalue condition often employed in the high-dimensional regression literature \cite{buhlmann_vandegeer}. While this assumption is not critical for our results, it is necessary for deriving the error bounds for the predictor.

\subsection{Main results}
\subsubsection{Slow rates}

We first provide non-asymptotic bounds on the excess quantile risk with minimal assumption.
\\
Put $ \tilde{\xi} = s^* \log ( n\sqrt{d} / s^*)
/\sqrt{n}  $.
\begin{theorem}
	\label{thm_main_2}
	Assume that Assumption \ref{assume_X_bounded} is satisfied and  that our loss function is bounded, i.e. $ \ell_\tau (y,x) \in [0,C] $.  Take $\lambda= \sqrt{n} $, $ \varsigma = ( C_{\rm x} n\sqrt{d})^{-1} $.  Then for all $ \theta^* $ such that $  \| \theta^*\|_1 \leq C_1 - 2d\varsigma $ we have that
	\begin{equation*} \mathbb{E}\,\mathbb{E}_{\theta\sim\hat{\rho}_{\lambda}} [R(\theta) ]- R^*
\leq 
\mathcal{C}_1
\,	\tilde{\xi} 
,
\end{equation*}	
and	 with probability at least $ 1-\varepsilon, \varepsilon\in (0,1) $ that
	\begin{equation*} \mathbb{E}_{\theta\sim\hat{\rho}_{\lambda}} [R(\theta) ]- R^*
	\leq 
\mathcal{C}'_1
[ \tilde{\xi} +	n^{-1/2} \log( 2 / \varepsilon) ]
 	,
	\end{equation*}
	for some constant $ \mathcal{C}_1 , \mathcal{C}'_1 > 0 $ depending only on $ C,C_1,C_{\rm x} $.
\end{theorem}

All proofs are detailed in Appendix \ref{sc_proofs}, where we utilize the `PAC-Bayesian bounds' approach from \cite{catonibook} as our main technical framework. For a comprehensive understanding of PAC-Bayes bounds and recent developments in the field, the reader is referred to the following sources: \cite{guedj2019primer,alquier2021user}. Originally proposed in \cite{STW,McA}, PAC-Bayesian bounds were designed to provide empirical bounds on the prediction risk of Bayesian estimators. Nonetheless, as highlighted in \cite{catoni2004statistical,catonibook}, this method also offers robust tools for deriving non-asymptotic bounds.

In Theorem \ref{thm_main_2}, we establish a relationship between the integrated prediction risk of our approach and the minimal achievable risk \( R^* \). The boundedness assumption is not too crucial to our technical proofs,  \cite{alquier2021user} suggests that alternative methods might allow for its relaxation. Although the excess risk error rate in Theorem \ref{thm_main_2} is slower by an order of \( n^{-1/2} \) compared to the results in Theorem \ref{thm_main1}, it is, to our knowledge, an entirely new finding. Furthermore, it requires only minimal assumptions, such as a bounded loss function and a bounded moment for the distribution of the covariate \(X\).

We will now obtain results for the mean estimator as a direct consequence of Theorem \ref{thm_main1}, utilizing Jensen's inequality and the convexity of the quantile loss. Therefore, the proof is not included.

\begin{cor}
	Assume Theorem \ref{thm_main_2} holds true, then we have that
	\begin{equation*}
\mathbb{E}\, R(\hat\theta) - R^*
\leq 
\mathcal{C}_1
\tilde{\xi}
,
\end{equation*}	
and	 with probability at least $ 1-\varepsilon, \varepsilon\in (0,1) $ that
	\begin{equation*}
 R(\hat\theta) - R^*
\leq 
\mathcal{C}'_1
[ \tilde{\xi}+	n^{-1/2} \log( 2 / \varepsilon) ]
,
\end{equation*}	
	for some constant $ \mathcal{C}_1 , \mathcal{C}_1' > 0 $ depending only on $ C,C_1,C_{\rm x} $.
\end{cor}

We now present a result concerning randomized estimators where \( \theta \) is sampled from \( \hat{\rho}_{\lambda} \). Here, it should be noted that `\textit{with probability at least}' in Proposition \ref{propo_slow} below means the probability evaluated with respect to the distribution $ \mathbf{P}^{\otimes n} $ of the data and the conditional Gibbs posterior distribution $ \hat{\rho}_{\lambda} $.

\begin{prop}
	\label{propo_slow}
	Assume that Assumption \ref{assume_X_bounded} is satisfied and  that our loss function is bounded, i.e. $ \ell_\tau (y,x) \in [0,C] $. Put $ \tilde{\xi} = s^* \log ( n\sqrt{d} / s^*)
/\sqrt{n}  $. Take $\lambda= \sqrt{n} $, $ \varsigma = ( C_{\rm x} n\sqrt{d})^{-1} $.  Then for all $ \theta^* $ such that $  \| \theta^*\|_1 \leq C_1 - 2d\varsigma $, we have with probability at least $ 1-\varepsilon, \varepsilon\in (0,1) $ that
\begin{equation*} 
R(\theta) - R^*
\leq 
\mathcal{C}''_1
[ \tilde{\xi} +	n^{-1/2} \log( 2 / \varepsilon) ]
,
\end{equation*}
for some constant $ \mathcal{C}''_1 > 0 $ depending only on $ C,C_1,C_{\rm x} $.
\end{prop}

\subsubsection{Fast rates}

Theorem \ref{thm_main_2} provides an oracle inequality that connects the integrated prediction risk of our method to the minimum achievable risk. However, these bounds can be improved by introducing additional assumptions.
\\
Put $ \tilde{\delta} = s^* \log ( n\sqrt{d} / s^*)
/ n $ and $ C_{_{K,C}} : = \max(2K,C) $.

\begin{theorem}
	\label{thm_main1}
	Assume that Assumption \ref{assume_X_bounded} and \ref{assum_bernstein} are satisfied and  that our loss function is bounded, i.e. $ \ell_\tau (y,x) \in [0,C] $. Take $\lambda= n/C_{_{K,C}}  $, $ \varsigma = ( C_{\rm x} n\sqrt{d})^{-1} $.  Then for all $ \theta^* $ such that $  \| \theta^*\|_1 \leq C_1 - 2d\varsigma $ we have:
	\begin{equation*}
\mathbb{E} \mathbb{E}_{\theta\sim\hat{\rho}_{\lambda}} [R(\theta) ]- R^*
\leq 
\mathcal{C}_2
\tilde{\delta} 
\end{equation*}
and	 with probability at least $ 1-\varepsilon, \varepsilon\in (0,1) $ that
\begin{equation*}
\mathbb{E}_{\theta\sim\hat{\rho}_{\lambda}} [R(\theta) ] - R^*
\leq 
\mathcal{C}_2'
[ \tilde{\delta}  +	\log( 2 / \varepsilon)/n ]
,
\end{equation*}	
for some constant $ \mathcal{C}_2 , \mathcal{C}_2' > 0 $ depending only on $ K, C,C_1,C_{\rm x} $.
\end{theorem}

In contrast to Theorem \ref{thm_main_2}, Theorem \ref{thm_main1} offers a bound that scales more rapidly, at \( 1/n \) rather than \( 1/\sqrt{n} \). Both results are presented in expectation and with high probability, facilitating a comparison between the out-of-sample error of our method and the optimal prediction error \( R^* \).

\begin{rmk}
The results presented in both Theorem \ref{thm_main1} and Theorem \ref{thm_main_2} are non-asymptotic and exhibit an adaptive nature, allowing for effective predictions without requiring knowledge of the sparsity level \( s^* \). 

In high-dimensional cases where \( n < d \), our prediction rates in Theorem \ref{thm_main1} follow the order \( s^* \log (d/s^*)/n \), aligning with the minimax-optimal rates  established for high-dimensional regression \cite{bellec2018slope,wang2019l_1}. It is important to note that \cite{chen2023sparse} achieved a sub-optimal prediction risk rate of \( s^* \log (d)/n \) using an \( \ell_0 \) penalization method.
\end{rmk}

As a direct corollary of Theorem \ref{thm_main1} by using Jensen's inequality, we now present the results for the mean estimator. 

\begin{cor}
	Assume Theorem \ref{thm_main1} holds true, then we have that
	\begin{equation*}
\mathbb{E} [R(\hat\theta) ]- R^*
\leq 
\mathcal{C}_2
\tilde{\delta} 
\end{equation*}
and	 with probability at least $ 1-\varepsilon, \varepsilon\in (0,1) $ that
\begin{equation*}
R(\hat\theta) - R^*
\leq 
\mathcal{C}_2'
[ \tilde{\delta} +	n^{-1} \log( 2 / \varepsilon) ]
,
\end{equation*}	
for some constant $ \mathcal{C}_2 , \mathcal{C}_2' > 0 $ depending only on $ K, C,C_1,C_{\rm x} $.
\end{cor}

Under Assumption \ref{assum_bernstein}, we are able to establish a fast excess risk bound, which provides a measure of how our predictor performs relative to the optimal predictor. This assumption not only facilitates a rapid assessment of the excess risk but also allows us to gain deeper insights into the relationship between our predictor and the optimal predictor, \( \theta^* \).

\begin{prop}
	\label{propo_fastrate}
Assume that Theorem \ref{thm_main1} holds, we deduce that
	\begin{equation*}
\mathbb{E}\, \mathbb{E}_{\theta\sim\hat{\rho}_{\lambda}}  \{ |  X^\top (\theta - \theta^*) |^2 \}
\leq 
C''_2
\tilde{\delta} 
\end{equation*}
and further assume that Assumption \ref{assum_eigen} satisfied, then
\begin{equation*}
\mathbb{E}\, \mathbb{E}_{\theta\sim\hat{\rho}_{\lambda}}  
	\|  \theta - \theta^* \|^2
\leq 
C'''_2
\frac{\tilde{\delta} 
}{ \kappa^2},
\end{equation*}
for some constant $ C''_2 > 0 $ depending only on $ K, C,C_1,C_{\rm x} $ and  $ C'''_2 > 0 $ depending only on $ \kappa, K, C,C_1,C_{\rm x} $.
\end{prop}

\begin{rmk}
Proposition \ref{propo_fastrate} offers non-asymptotic bounds on the mean-square error of our predictors. It is worth noting that our rate \( \tilde{\delta} = s^* \log (d/s^*)/n \) is minimax-optimal, as established by \cite{wang2019l_1}. In contrast, the \( \ell_0 \) approach presented in \cite{chen2023sparse} achieves a sub-optimal rate of \( s^* \log (d)/n \), which is similar as in \cite{belloni2011ell_1}.
\end{rmk}

This section outlines the chosen values for the tuning parameters \( \lambda \) and \( \varsigma \) in our proposed method, based on theoretical prediction error rates. While these values offer a foundation, they may not be optimal in practical settings. Practitioners can utilize, for example, cross-validation to further refine these parameters for specific applications. Nonetheless, the theoretical values presented here offer a valuable benchmark for gauging the appropriate scale of the tuning parameters in real-world scenarios.

\section{Simulation studies}
\label{sc_numerical}

\subsection{Implementation}

The gradient of the logarithmic of the Gibbs posterior in \eqref{eq_Gibbs_poste} can be obtained as
$
\nabla \log \hat{\rho}_{\lambda} (\theta) \propto - \lambda\nabla  r_n (\theta) + \nabla  \log \pi (\theta)
$
and $ \nabla  r_n (\theta) = \sum_{i=1}^{n} X_i ( \tau \mathbf{1}_{Y_i > X_i^\top\theta } +  (\tau-1) \mathbf{1}_{Y_i \leq X_i^\top\theta } ) $. Thus, our approach is implemented using a constant step-size unadjusted Langevin Monte Carlo (LMC) algorithm, \cite{durmus2019high}. The algorithm initializes with an initial point \( \theta_0 \) and iterates according to the update rule:
\begin{equation}
\label{langevinMC}
\theta_{s+1} = \theta_{s} - \eta \nabla \log \hat{\rho}_{\lambda}(\theta_s) + \sqrt{2\eta} W_s \qquad s=0,1,\ldots
\end{equation}
Here, \( \eta > 0 \) is the step-size, and \( W_0, W_1, \ldots \) are independent random vectors with i.i.d. standard Gaussian components. It is crucial to select an appropriate step-size \( \eta \), as an excessively small value could lead to instability in the sum, a problem highlighted in \cite{roberts2002langevin}.

To ensure convergence to the desired distribution, we also explore the Metropolis-adjusted Langevin algorithm (MALA), which integrates a Metropolis–Hastings (MH) correction into the process. Nonetheless, this method can lead to slower performance due to the extra acceptance/rejection step that must be performed at every iteration.

\subsubsection*{Compared methods} 

Our proposed methods, LMC and MALA, are compared against two leading methods: one from the frequentist framework and one from the Bayesian framework. The first comparison is with a lasso-type penalized quantile regression approach \cite{yi2017semismooth}, denoted `Lasso', implemented in the \texttt{R} package `\texttt{hqreg}' \cite{hqreg}. The second comparison is with a newly introduced Bayesian quantile regression method using the Horseshoe prior in \cite{kohns2024horseshoe} with the corresponding \texttt{R} code available, this method is denoted by `Horshoe'.

\subsection{Simulation setup}
The data in the simulation studies are generated by
\begin{align*}
y_i = x_i^\top \theta^* + u_i, \quad i = 1, \ldots n, 	
\end{align*}
where $ \theta^* \in \mathbb{R}^d $ and the noise random part $ u_i $'s $ \tau $th quantile equal to 0. We examine two configurations for the dimensions $(n,d)$, where $(n,d) = (50,100)$ represents a small-scale scenario, and $(n,d) = (200,400)$ represents a larger-scale scenario. For each configuration, we evaluate two scenarios for the sparsity level $s^*$, which denotes the number of non-zero elements in the parameter vector $ \theta^* $. Specifically, we consider $s^* = 5$ and $s^* = n/2$. In all cases, the non-zero elements of the true parameter vector $ \theta^* $ are independently drawn from a standard normal distribution $ \mathcal{N}(0,1) $. The elements of the covariate matrix $X$ are also generated independently from $ \mathcal{N}(0,1) $. These considerations yield four distinct settings, corresponding to the four tables presented. Within  each setting, we consider three different choices for the distribution of $ u_i $'s: 
\begin{itemize}
	\item The first choice is a normal distribution $ \mathcal{N} (\mu,\sigma^2) $ with $ \mu = 0, \sigma^2 = 9 $.
	\item The second choice is a Cauchy distribution $ {\rm Cau }(0,1) $.
	\item The third one is a Student t-distribution with 3 degree of freedom and scaled by a factor 2.
\end{itemize}

The methods are evaluated using the following mean prediction error (mpe) and mean squared error (mse): 
$$
 {\rm mpe}: = n^{-1} \sum_{i=1}^{n}
\ell_\tau (Y_i , X_i^\top \theta) 
, \quad
 {\rm mse}: = d^{-1}
\| \widehat{\theta} -\theta^*\|_2^2 ,
,
$$
where $ \widehat{\theta} $ is one of the considered methods.
Each simulation setting is repeated 100 times and we report the averaged results for different errors. The results of the simulations study are detailed in Table \ref{tb_01}, \ref{tb_02}, \ref{tb_03}, \ref{tb_04}  and the values within parentheses indicate the standard deviation associated with the error.

The LMC, MALA are run with 30000 iterations and the first 500 steps are discarded as burn-in period. The LMC, MALA are initialized at the Lasso estimate. We set the values of tuning parameters $ \lambda  $ and $ \varsigma  $
to 1 for all scenarios. It is important to acknowledge that a better approach could be to tune these parameters using cross validation, which could lead to improved results. The Gibbs sampler for Bayesian approach with Horseshoe prior is run with 1500 iterations and the first 500 steps are discarded as burn-in. The quantile Lasso method is run with default options and that 5-fold cross validation is used to select the tuning parameter.

\subsection{Simulation results}

As shown in Tables \ref{tb_01}, \ref{tb_02}, \ref{tb_03}, and \ref{tb_04}, all the methods considered generally perform comparably, with no single method consistently outperforming the others across all simulation setups. In other words, our proposed method is competitive with the state-of-the-art approaches from both frequentist and Bayesian frameworks. The detailed simulation results are as follows.

In the case of small sample sizes and very sparse settings, as shown in Table \ref{tb_01}, the Horseshoe method consistently performs well in terms of prediction error, often leading the pack. In contrast, Lasso tends to have the worst predictive performance in this area, struggling significantly in these settings. Our proposed method, implemented via MALA, usually ranks just below the Horseshoe method, demonstrating strong predictive capabilities. Although the LMC-based approach is not as effective as MALA, it still outperforms Lasso, underscoring its effectiveness in this specific setting. When it comes to estimation error, however, the ranking shifts. Lasso emerges as the best performer, suggesting its strength in accurately estimating parameters, with Horseshoe taking the second place. This indicates that while Lasso may not excel in prediction, it offers advantages in estimation accuracy, particularly in very sparse settings.

\begin{table}[!h]
	\centering
	\caption{ Simulation results. $ n = 50, p = 100, s^* = 5 $}
	\begin{tabular}{ l l l cccc  }
		\hline \hline
 noise & quantile & error  & LMC & MALA & Lasso & Horshoe
		\\
		\hline
  $ \mathcal{N} (0,3^2) $ &
$ \tau =0.1 $ & pred.error & 0.526 (0.148) & 0.272 (0.038) & 0.759 (0.523) & 0.116 (0.023)
		\\
 & & est.error		& 0.134 (0.026) & 0.180 (0.035) & 0.046 (0.027) & 0.075 (0.029)
		\\
  &
$ \tau =0.5 $ & pred.error & 0.867 (0.154) & 0.657 (0.113) & 1.003 (0.372) & 0.412 (0.091)
		\\
& & est.error
  & 0.112 (0.027) & 0.146 (0.025) & 0.041 (0.021) & 0.053 (0.020)
		\\
  &
$ \tau =0.9 $ & pred.error & 0.523 (0.157) & 0.407 (0.123) & 0.852 (0.566) & 0.118 (0.032)
		\\
& & est.error &
 0.131 (0.026) & 0.143 (0.028) & 0.043 (0.027) & 0.073 (0.026)
		\\
		\hline
		\\
   $ {\rm Cau }(0,1) $ &
$ \tau =0.1 $ 
& pred.error & 2.298 (4.254) & 1.898 (4.172) & 2.605 (4.379) & 2.044 (4.272)
		\\
& & mse
& 0.123 (0.029) & 0.213 (0.052) & 0.038 (0.024) & 0.045 (0.039)
\\
 &
$ \tau =0.5 $  & pred.error & 3.527 (9.489) & 3.265 (9.497) & 3.514 (9.534) & 3.165 (9.450)
		\\
& & mse
& 0.113 (0.026) & 0.161 (0.038) & 0.048 (0.030) & 0.026 (0.021)
\\
 &
$ \tau =0.9 $ & pred.error & 2.142 (3.284) & 2.006 (3.281) & 2.484 (3.289) & 1.837 (3.323)
		\\
&  & mse
& 0.127 (0.027) & 0.150 (0.035) & 0.045 (0.028) & 0.057 (0.056)
		\\
		\hline
		\\
  $ 2\cdot t_3 $ & $ \tau =0.1 $ & pred.error &
 0.530 (0.182) & 0.297 (0.091) & 0.765 (0.468) & 0.159 (0.117)
 		\\
 & & mse
 & 0.128 (0.030) & 0.189 (0.043) & 0.043 (0.028) & 0.068 (0.031)
 \\
  &
 $ \tau =0.5 $  & pred.error &
 0.884 (0.189) & 0.666 (0.155) & 0.896 (0.368) & 0.453 (0.155)
 		\\
 & & mse
 & 0.105 (0.021) & 0.134 (0.027) & 0.035 (0.022) & 0.037 (0.017)
 \\
  &
 $ \tau =0.9 $ & pred.error &
0.601 (0.253) & 0.486 (0.227) & 0.898 (0.521) & 0.188 (0.144)
 		\\
 & & mse
 & 0.130 (0.031) & 0.145 (0.034) & 0.047 (0.030) & 0.067 (0.043)
		\\
		\hline
\hline	
\end{tabular}
\label{tb_01}
\end{table}

\begin{table}[!h]
\centering
\caption{Simulation results. $ n = 50, p = 100,s^* = 25  $}
\begin{tabular}{ l l l cccc  }
\hline \hline
 noise & quantile & error  & LMC & MALA & Lasso & Horshoe
\\ \hline
 $ \mathcal{N} (0,3^2) $ & $ \tau =0.1 $ & mpe &
0.745 (0.282) & 0.339 (0.061) & 0.835 (0.808) & 0.143 (0.036)
\\
&& mse &
0.287 (0.065) & 0.334 (0.072) & 0.234 (0.064) & 0.190 (0.067)
\\
 &
$ \tau =0.5 $  &  mpe &
1.159 (0.330) & 0.871 (0.210) & 0.915 (0.627) & 0.586 (0.147)
\\
&& mse &
0.247 (0.062) & 0.262 (0.061) & 0.211 (0.075) & 0.154 (0.053)
\\
 &
$ \tau =0.9 $ &  mpe &
0.716 (0.301) & 0.618 (0.238) & 0.735 (0.793) & 0.145 (0.033)
\\
&& mse &
0.281 (0.068) & 0.283 (0.067) & 0.230 (0.074) & 0.187 (0.065)
		\\
\hline
		\\
  $ {\rm Cau }(0,1) $ &
$ \tau =0.1 $ &  mpe &
2.764 (6.387) & 2.148 (6.278) & 2.989 (6.345) & 2.075 (6.353)
\\
&& mse &
0.293 (0.061) & 0.374 (0.076) & 0.243 (0.058) & 0.205 (0.081)
\\
 &
$ \tau =0.5 $  &  mpe &
13.40 (108.5) & 13.01 (108.5) & 13.17 (108.6) & 12.73 (108.5)
\\
&& mse &
0.264 (0.063) & 0.292 (0.069) & 0.240 (0.071) & 0.156 (0.078)
\\
 &
$ \tau =0.9 $ &  mpe &
2.856 (4.797) & 2.672 (4.769) & 3.065 (4.872) & 2.148 (4.802)
\\
&& mse &
0.278 (0.062) & 0.285 (0.064) & 0.234 (0.066) & 0.203 (0.106)
		\\
		\hline
		\\
 $ 2\cdot t_3 $ & $ \tau =0.1 $ &  mpe &
  0.799 (0.334) & 0.378 (0.135) & 0.877 (0.812) & 0.193 (0.143)
\\
&& mse &
0.282 (0.079) & 0.335 (0.076) & 0.229 (0.077) & 0.175 (0.068)
\\
 &
$ \tau =0.5 $  &   mpe &
1.129 (0.307) & 0.851 (0.199) & 0.940 (0.627) & 0.602 (0.163)
\\
&& mse &
0.245 (0.061) & 0.267 (0.062) & 0.207 (0.067) & 0.154 (0.058)
\\
&
$ \tau =0.9 $ &  mpe &
 0.813 (0.297) &
0.667 (0.250) & 0.883 (0.866) & 0.173 (0.096)
\\
&& mse &
0.297 (0.071) & 0.290 (0.076) & 0.245 (0.077) & 0.192 (0.064)
		\\
		\hline
	\hline	
	\end{tabular}
\label{tb_02}

	\caption{Simulation results. $ n = 200, p = 400, s^* = 5 $}
	\begin{tabular}{ l l l cccc  }
		\hline \hline
 noise & quantile & error & LMC & MALA & Lasso & Horshoe
		\\
		\hline
 $ \mathcal{N} (0,3^2) $ & $ \tau =0.1 $ &   mpe &
  0.619 (0.088)
	& 0.273 (0.020) & 1.042 (0.294) & 0.324 (0.077)
		\\
& &	 mse 	 & 0.026 (0.004)
	& 0.043 (0.005) & 0.009 (0.005) & 0.009 (0.002)
		\\
 & $ \tau =0.5 $  &   mpe &
  0.926 (0.070) & 0.681 (0.055)
 & 1.102 (0.199) & 0.692 (0.064)
		\\
		& &	 mse 
			 & 0.021 (0.003)
& 0.033 (0.003) & 0.006 (0.003) & 0.005 (0.002)
		\\
 &	$ \tau =0.9 $ &   mpe &
  0.609 (0.091) & 0.451 (0.069) &
1.063 (0.322) & 0.317 (0.070)
		\\
& &	 mse 	 & 0.025 (0.003)
		& 0.035 (0.004) & 0.008 (0.004) & 0.009 (0.002)
				\\
		\hline
		\\
  $ {\rm Cau }(0,1) $ &
		$ \tau =0.1 $ &   mpe &
	 5.791 (22.25) & 5.272 (22.23) & 6.083 (22.23) & 5.753 (22.18)
		\\
		& &	 mse 	 & 0.026 (0.005)
		& 0.050 (0.006) & 0.011 (0.007) & 0.002 (0.002)
		\\
 &	$ \tau =0.5 $  &   mpe &
  4.857 (20.47) &
4.569 (20.46) & 4.815 (20.46) & 4.628 (20.40)
		\\
		& &		 mse  & 0.024 (0.005)
		& 0.038 (0.005) & 0.011 (0.007) & 0.001 (0.001)
		\\
& 	$ \tau =0.9 $ &   mpe &
 2.259 (2.101) &
2.065 (2.087) & 2.540 (2.083) & 2.243 (2.080)
		\\
		& &	 mse 	 & 0.026 (0.008)
		& 0.037 (0.006) & 0.011 (0.008) & 0.002 (0.001)
				\\
		\hline
		\\
 $ 2\cdot t_3 $ & $ \tau =0.1 $ &  mpe &
  0.607 (0.119) & 0.288 (0.044) & 0.990 (0.254) & 0.415 (0.116)
		\\
		& &	 mse 	 & 0.024 (0.003)
		& 0.041 (0.004) & 0.007 (0.003) & 0.006 (0.001)
		\\
& 	$ \tau =0.5 $  &   mpe &
 0.928 (0.107) & 0.706 (0.078) & 1.046 (0.165) & 0.728 (0.093)
		\\
		& &	 mse 	 & 0.019 (0.002)
		& 0.032 (0.003) & 0.005 (0.003) & 0.003 (0.001)
		\\
& 	$ \tau =0.9 $ &   mpe &
 0.607 (0.111) & 0.455 (0.081) & 0.980 (0.244) & 0.406 (0.121)
		\\
& &	 mse 	 & 0.024 (0.003)
& 0.034 (0.003) & 0.008 (0.004) & 0.005 (0.002)
\\
		\hline
\hline	
\end{tabular}
\label{tb_03}
\end{table}

\begin{table}[!h]
\centering
\caption{Simulation results. $ n = 200, p = 400, s^* = 100 $}
\begin{tabular}{ l l l cccc  }
\hline \hline
 noise & quantile & error  & LMC & MALA & Lasso & Horshoe
\\ \hline
 $ \mathcal{N} (0,3^2) $ &
		$ \tau =0.1 $ &    mpe &
		 1.223 (0.730) & 0.648 (0.233) & 0.584 (0.911) & 1.211 (0.202)
		\\
		& &	 mse 	 & 0.194 (0.041)
		& 0.189 (0.038) & 0.193 (0.049) & 0.143 (0.029)
		\\
& 	$ \tau =0.5 $  &    mpe & 1.120 (0.347) & 1.049 (0.282) & 0.407 (0.397) & 2.253 (0.252)
		\\
		& &	 mse 	 & 0.132 (0.028)
		& 0.128 (0.027) & 0.126 (0.031) & 0.188 (0.035)
		\\
& $ \tau =0.9 $ &   mpe & 1.147 (0.683) & 1.056 (0.457) & 0.441 (0.654) & 1.232 (0.213)
		\\
& &	 mse 	 & 0.191 (0.037)
& 0.184 (0.036) & 0.189 (0.046) & 0.146 (0.032)
				\\
		\hline
		\\
  $ {\rm Cau }(0,1) $ &
		$ \tau =0.1 $ &   mpe &
		 5.090 (4.444) & 3.526 (4.358) & 3.810 (4.662) & 3.588 (4.222)
		\\
		& &	 mse 	 & 0.228 (0.034)
		& 0.212 (0.030) & 0.239 (0.038) & 0.143 (0.028)
		\\
		& 
		$ \tau =0.5 $  &   mpe & 
4.467 (4.659) & 3.874 (4.514) & 3.077 (4.408) & 4.317 (4.264)
		\\
		& &	 mse 	 & 0.195 (0.049)
		& 0.183 (0.042) & 0.209 (0.058) & 0.185 (0.036)
		\\
		& 
$ \tau =0.9 $ &   mpe & 5.030 (8.268) & 4.629 (8.239) & 4.294 (8.427) & 4.127 (8.059)
		\\
& &	 mse 	 & 0.231 (0.042)
& 0.222 (0.042) & 0.246 (0.048) & 0.150 (0.036)
				\\
		\hline
		\\
$ 2\cdot t_3 $ & $ \tau =0.1 $ &   mpe &
 1.642 (1.018) & 0.821 (0.367) & 0.780 (1.116) & 1.276 (0.231)
		\\
		& &	 mse 	 & 0.194 (0.047)
		& 0.188 (0.040) & 0.196 (0.053) & 0.141 (0.032)
		\\
		& 
		$ \tau =0.5 $  &   mpe & 1.239 (0.399) & 1.142 (0.285) & 0.551 (0.455) & 2.289 (0.247)
		\\
		& &		 mse  & 0.133 (0.030)
		& 0.129 (0.029) & 0.128 (0.034) & 0.186 (0.033)
		\\
& 	$ \tau =0.9 $ &   mpe & 1.331 (0.740) & 1.293 (0.599) & 0.646 (0.892) & 1.263 (0.201)
		\\
		& &	 mse 	 & 0.198 (0.037)
		& 0.192 (0.037) & 0.199 (0.046) & 0.147 (0.034)
		\\
		\hline
		\hline
	\end{tabular}
\label{tb_04}
\end{table}

The situation shifts somewhat in the case of small sample sizes but less sparse settings, where \( s^* = n/2 = 25 \), as illustrated in Table \ref{tb_02}. In this scenario, one of the most remarkable observations is that the estimation errors are quite comparable across all the methods considered, with the Horseshoe method emerging as the leading approach in this regard. While Lasso continues to struggle, showing the highest prediction error and failing to perform well in this setting, Horseshoe demonstrates strong predictive capabilities, consistently delivering lower prediction errors. Our proposed methods, implemented via MALA and LMC, also perform commendably in this scenario, ranking as the second-best options. Although MALA and LMC do not surpass Horseshoe, they still outperform Lasso, indicating their effectiveness in this specific setting.

Results in Table \ref{tb_03}, which address the case of higher dimensions and a very sparse setting, highlight the superiority of our proposed methods. Notably, MALA frequently achieves the smallest prediction errors, in some instances outperforming Lasso by a factor of 2 or 3. The Horseshoe method proves to be consistently stable, ranking as the second-best in terms of predictive performance, while Lasso continues to exhibit the highest prediction errors, struggling in this context. When considering estimation error, both Lasso and Horseshoe demonstrate excellent performance, with very low estimation errors. Although the estimation errors for our methods are also quite small, they remain slightly higher than those of Lasso and Horseshoe. This indicates that while our methods excel in prediction, particularly in challenging settings, they still lag behind Lasso and Horseshoe when it comes to estimation accuracy.

Increasing the sparsity to \( s^* = n/2 = 100 \), as shown in Table \ref{tb_04}, reveals some promising developments for the Lasso method. Specifically, in this scenario, Lasso frequently achieves the smallest prediction errors, marking a significant improvement in its performance. Our MALA method typically ranks as the second-best, consistently delivering strong predictive results. Meanwhile, Horseshoe, which previously performed well, now stands in third place, showing predictive performance comparable to that of the LMC method. Regarding estimation error, Horseshoe and our MALA method emerge as the top contenders, demonstrating comparable and highly competitive performance. Interestingly, similar to the results observed in Table \ref{tb_02}, Lasso's performance in terms of estimation error declines as the sparsity increases. It no longer excels in estimation accuracy and is now on par with the LMC method. This shift underscores the complexity of balancing predictive and estimation accuracy in varying levels of sparsity.

\section{Applications}
\label{sc_realdata}
\subsection{Predicting fat content of ground pork}

We first illustrate the application of our proposed method using the `\texttt{meatspec}' dataset. This dataset includes measurements from finely chopped pure meat samples, totaling 215 observations. For each sample, both fat content and a 100-channel spectrum of absorbances were recorded \cite{borggaard1992optimal}. Given that determining fat content through analytical chemistry is labor-intensive, our goal is to develop a model that can predict fat content in new samples based on the 100 absorbance readings, which are more straightforward to collect. Our analysis utilizes the 215 samples available in the \texttt{R} package `\texttt{faraway}' \cite{faraway2016faraway}. The fat content data undergoes a log transformation before being scaled and centered  to have a mean of zero and a standard deviation of one. Similarly, the channel spectrum is also adjusted by scaling and centering so that it has a zero mean and a standard deviation of one.

For comparison, we randomly select 151 out of the 215 samples as training data, with the remaining 64 samples used as test data, resulting in a roughly 70/30 percent of the data split. The methods are executed on the training set, and their predictive accuracy is subsequently evaluated on the test set. This procedure is repeated 100 times with different random splits of training and testing data each time. Repeating and averaging the process across multiple iterations provides a more stable and precise assessment of the methods, helping to mitigate the impact of data variability and enhancing our understanding of their performance. The results are summarized in Table \ref{tb_meat_fat}.

\begin{table}[!h]
	\centering
	\caption{Means (and standard deviations) quantile prediction errors for the ground pork data}
	\begin{tabular}{  l  cccc  }
		\hline \hline
 quantile & LMC & MALA & Lasso & Horshoe
		\\ 
		\hline
$ \tau =0.1 $ &  0.2205 (0.0447) 
	 & 0.2228 (0.0452) & 0.2262 (0.0440) & 0.2058 (0.0439)
		\\
	$ \tau =0.5 $  &  0.2170 (0.0227) 
	& 0.2293 (0.0251) & 0.2276 (0.0210) & 0.2163 (0.0234)
		\\
	 $ \tau =0.9 $ &  0.2192 (0.0226) 
	 & 0.2526 (0.0425) & 0.2284 (0.0241) & 0.2080 (0.0213)
		\\
		\hline
		\hline
	\end{tabular}
	\label{tb_meat_fat}
\end{table}

The results presented in Table \ref{tb_meat_fat} indicate that all the methods evaluated perform well, demonstrating similar predictive accuracy and behavior. Among the methods, the Bayesian Horseshoe approach emerges as the top performer, though its advantage over the other techniques is minimal. Additionally, it is worth noting that the LMC method shows a slight edge over MALA, though this difference is also quite marginal. Overall, the performance across the different methods is comparable, with only subtle distinctions in their effectiveness.

\subsection{Predicting TRIM32 expression using gene expression data}

In this application, we conduct an analysis of quantile prediction using high-dimensional genomics data from \cite{scheetz2006regulation}. The study by \cite{scheetz2006regulation} involved analyzing RNA from the eyes of 120 twelve-week-old male rats, using 31,042 different probe sets. Our focus is on modeling the expression of the gene TRIM32, as it was identified by \cite{chiang2006homozygosity} as a gene associated with Bardet-Biedl syndrome, a condition that includes retinal degeneration among its symptoms. Since \cite{scheetz2006regulation} observed that many probes were not expressed in the eye, we follow the approach of \cite{huang2010variable} and \cite{sherwood2022additive}, limiting our analysis to the 500 genes with the highest absolute Pearson correlation with TRIM32 expression. The data for this analysis is available from the \texttt{R} package `\texttt{abess}', \cite{zhu2022abess}.

To assess the methods, we randomly allocate 84 of the 120 samples for training and the remaining 36 for testing, maintaining an approximate 70/30 percent of the data split. The methods are executed using the training set, and their prediction accuracy is evaluated on the test set. This procedure is repeated 100 times, each with a different random partition of the data. The outcomes of these iterations are displayed in Table \ref{tb_trim32}.

\begin{table}[!h]
	\centering
	\caption{Means (and standard deviations) quantile prediction errors for the \texttt{trim32} data}
	\begin{tabular}{  l  cccc  }
		\hline \hline
		quantile & LMC & MALA & Lasso & Horshoe
		\\ 
		\hline
$ \tau =0.1 $ & 0.0213 (0.0048) & 0.0804 (0.0124) & 0.0118 (0.0058) & 0.0184 (0.0093)
		\\
$ \tau =0.5 $  & 0.0454 (0.0073) & 0.0727 (0.0108) & 0.0431 (0.0050) & 0.0268 (0.0098)
		\\
$ \tau =0.9 $ & 0.0208 (0.0024) & 0.0455 (0.0134) & 0.0177 (0.0027) & 0.0161 (0.0081)
		\\
		\hline
		\hline
	\end{tabular}
	\label{tb_trim32}
\end{table}

In contrast to the findings presented in Table \ref{tb_meat_fat}, the results shown in Table \ref{tb_trim32} reveal a different scenario. For the \texttt{trim32} dataset, the Bayesian Horseshoe method stands out as the most effective, securing the top position. The Lasso method follows closely as the second-best performer. Our proposed method, which employs LMC, trails Lasso by a small margin. Although MALA produces relatively low prediction errors, they are still higher compared to the other methods considered.

\section{Conclusion}
\label{sc_conclusion}
This paper presents a novel probabilistic machine learning approach for high-dimensional quantile prediction, employing PAC-Bayes bounds to derive non-asymptotic oracle inequalities for the method. These inequalities demonstrate that our technique achieves minimax-optimal prediction error and adapts to the unknown sparsity of the underlying model. In addition to its solid theoretical underpinnings, our approach provides practical insights for implementation. By using a scaled Student-t prior, we enable the use of Langevin Monte Carlo (LMC) for efficient computation. The effectiveness of our method is demonstrated through simulations and real data, where it is compared with two established frequentist and Bayesian techniques.

An area that has not been addressed in this work is variable selection, which could be an interesting and valuable direction for future research. Variable selection plays a crucial role in improving model interpretability and reducing computational complexity, especially in high-dimensional settings where irrelevant variables can obscure the true signal. Investigating how our probabilistic machine learning method can be extended to incorporate variable selection techniques would not only enhance its practical applicability but also potentially lead to more efficient and interpretable models.

While our proposed method demonstrates  effectiveness in both simulated and real datasets, we acknowledge that Langevin Monte Carlo (LMC) algorithms may encounter scalability challenges in the context of large-scale data. 
To overcome these limitations, variational inference (VI) has been identified as a promising alternative to traditional Markov chain Monte Carlo techniques. VI offers a computational optimization-based approach that scales more efficiently with large datasets, making it an attractive option for extending the scalability of our method. The connection between the PAC-Bayesian framework, which underpins our theoretical results, and variational inference has been explored in detail in \cite{alquier2016}, highlighting potential avenues for integrating these approaches.

\subsection*{Acknowledgments}
The author acknowledges support from the Norwegian Research Council, grant number 309960 through the Centre for Geophysical Forecasting at NTNU.

\subsection*{Conflicts of interest/Competing interests}
The author declares no potential conflict of interests.

\appendix
\section{Proofs}
\label{sc_proofs}

Let \( \mathcal{P}(\Theta) \) denote the set of all probability measures on \(\Theta\).
\subsection{Proof for slow rate}

\begin{proof}[\bf Proof for Theorem \ref{thm_main_2}]
		\text{}
	\\
\noindent \textit{Step 1:}
\\	
Fix $\theta\in\Theta$, for $ \ell_{i}(\theta)= \ell_\tau (Y_i , X_i^\top \theta)  $, and apply Hoeffding's Lemma \ref{lemma:hoeffding} with $U_i= \ell_i(\theta^*) - \ell_{i}(\theta)  $, one notes that $\mathbb{E} U_i = R(\theta^*)-R(\theta)$. We obtain for any $t>0$ that
$
	\mathbb{E}  {\rm e}^{ t n 
	[R(\theta) - R(\theta^*) -r(\theta) + r(\theta^*)] }  
	\leq 
	{\rm e}^{\frac{n t^2 C^2}{ 8}}.
$
	We put $t=\lambda/n$, which gives:
	\begin{equation*}
	\mathbb{E}  {\rm e}^{ \lambda 	[R(\theta) - R(\theta^*) -r(\theta) + r(\theta^*)]  } 
	\leq 
	{\rm e}^{\frac{\lambda^2 C^2}{ 8 n}}.
	\end{equation*}
Integrate this bound with respect to $\pi$ and use Fubini's theorem to obtain:
	\begin{equation}
	\label{eq_bound_for_randomized_slow}
	\mathbb{E}  \mathbb{E}_{\theta\sim\pi} \left[ {\rm e}^{ \lambda 	
[R(\theta) - R(\theta^*) -r(\theta) + r(\theta^*)] } \right] 
	\leq 
	{\rm e}^{\frac{\lambda^2 C^2}{ 8 n}}
	\end{equation}
	and we apply Lemma~\ref{lemma:dv} to get:
	\begin{equation*}
	\mathbb{E} \left[ {\rm e}^{ \sup_{\rho\in\mathcal{P}(\Theta)} \lambda \mathbb{E}_{\theta\sim\rho}
[R(\theta) - R(\theta^*) -r(\theta) + r(\theta^*)] -	\mathcal{K}(\rho\|\pi) } \right] 
	\leq 
	{\rm e}^{\frac{\lambda^2 C^2}{ 8 n}}.
	\end{equation*} 
	Rearranging terms:
	\begin{equation}
	\label{eq_to_provein_probabiliy}
	\mathbb{E} \left[ {\rm e}^{ \sup_{\rho\in\mathcal{P}(\Theta)} \lambda \mathbb{E}_{\theta\sim\rho}
[R(\theta) - R(\theta^*) -r(\theta) + r(\theta^*)] -	\mathcal{K}(\rho\|\pi)-\frac{\lambda^2 C^2}{ 8 n} } \right] 
	\leq 
	1
	.
	\end{equation}
Uses Chernoff bound  to get
	\begin{equation*}
	\mathbb{P} \left[ \sup_{\rho\in\mathcal{P}(\Theta)} \lambda \mathbb{E}_{\theta\sim\rho}
[R(\theta) - R(\theta^*) -r(\theta) + r(\theta^*)]
	-	\mathcal{K}(\rho\|\pi)-\frac{\lambda^2 C^2}{ 8 n}  > \log\frac{1}{\varepsilon} \right] 
	\leq 
	\varepsilon.
	\end{equation*} 
	Take the complement and rearranging terms give, with probability at least $ 1- \varepsilon $
	\begin{equation*}
 \mathbb{E}_{\theta\sim\rho}[ R(\theta) - R(\theta^*)] 
 \leq
 \mathbb{E}_{\theta\sim\rho}[ r(\theta) -  r(\theta^*)] +  \frac{\lambda C^2}{8n} + \frac{	\mathcal{K}(\rho\|\pi) + \log\frac{1}{\varepsilon}}{\lambda} 
.
	\end{equation*}
	using Lemma \ref{lemma:dv} again, we get that with probability at least $ 1- \varepsilon $
	\begin{equation}
	\label{eq_pacb_slow_1}
\mathbb{E}_{\theta \sim \hat{\rho}_\lambda}[ R(\theta) - R(\theta^*)] 
\leq
 \inf_{\rho\in\mathcal{P}(\Theta)}  
 \left\{ 
\mathbb{E}_{\theta\sim\rho}[ r(\theta) -  r(\theta^*)] +  \frac{\lambda C^2}{8n} + \frac{	\mathcal{K}(\rho\|\pi) + \log\frac{1}{\varepsilon}}{\lambda} 
\right\} 
.
\end{equation}	
Now, with exactly the same argument but with an application to $ -U_i $ rather than $ U_i $, one obtains that  with probability at least $ 1- \varepsilon $
\begin{equation}
	\label{eq_pacb_slow_2}
\mathbb{E}_{\theta\sim\rho}[ r(\theta) -  r(\theta^*)]
\leq
\mathbb{E}_{\theta\sim\rho}[ R(\theta) - R(\theta^*)] 
 +  \frac{\lambda C^2}{8n} + \frac{	\mathcal{K}(\rho\|\pi) + \log\frac{1}{\varepsilon}}{\lambda} 
.
\end{equation}
Combining \eqref{eq_pacb_slow_1} and \eqref{eq_pacb_slow_2}, we get that
with probability at least $ 1- 2\varepsilon $
	\begin{equation}
\label{eq_pacb_slow_3}
\mathbb{E}_{\theta \sim \hat{\rho}_\lambda}[ R(\theta) - R(\theta^*)] 
\leq
 \inf_{\rho\in\mathcal{P}(\Theta)}  \left\{ 
\mathbb{E}_{\theta\sim\rho}
[ R(\theta) - R(\theta^*)] 
+  2\frac{\lambda C^2}{8n} 
+ 2 \frac{	\mathcal{K} (\rho\|\pi) + \log\frac{1}{\varepsilon}}{\lambda} 
\right\} 
.
\end{equation}

To obtain bounds in expectation, from \eqref{eq_to_provein_probabiliy}, in particular for $\rho=\hat{\rho}_\lambda$, we use Jensen's inequality and rearranging terms:
\begin{equation*}
\lambda \left\{ \mathbb{E} \mathbb{E}_{\theta\sim\hat{\rho}_{\lambda}} [R(\theta) ]-R(\theta^*) \right\}
\leq 
\mathbb{E} \left[ \lambda [ \mathbb{E}_{\theta\sim\hat{\rho}_{\lambda} } [r(\theta) ]-r(\theta^*)] + 	
\mathcal{K}(\hat{\rho}_{\lambda}\|\pi)
+
\frac{\lambda^2 C^2}{ 8 n}
\right].
\end{equation*}
For $ \lambda >0 $ and note that $\hat{\rho}_\lambda$ minimizes the quantity in the expectation in the right-hand side (Lemma \ref{lemma:dv}), this can be rewritten:
\begin{align}
 \mathbb{E} \mathbb{E}_{\theta\sim\hat{\rho}_{\lambda}} [R(\theta) ]-R(\theta^*)
& \leq 
\mathbb{E}  \inf_{\rho\in\mathcal{P}(\Theta)}  
\left[   \mathbb{E}_{\theta\sim \rho} [r(\theta) ]-r(\theta^*) + 	
 \frac{\mathcal{K}(\rho\|\pi)}{\lambda}
+
\frac{\lambda C^2}{ 8 n}
\right] \nonumber
\\
&\leq 
 \inf_{\rho\in\mathcal{P}(\Theta)}  
\left[   \mathbb{E}_{\theta\sim \rho} 
\mathbb{E}  [r(\theta) -r(\theta^*)] + 	
\frac{\mathcal{K}(\rho\|\pi)}{\lambda}
+
\frac{\lambda C^2}{ 8 n}
\right] \nonumber
\\
&\leq 
\inf_{\rho\in\mathcal{P}(\Theta)}  
\left[   \mathbb{E}_{\theta\sim \rho} 
  [R(\theta)] - R ( \theta^*) + 	
\frac{\mathcal{K}(\rho\|\pi)}{\lambda}
+
\frac{\lambda C^2}{ 8 n}
\right]
\label{eq_bound_in_EXPECT_slow}
.
\end{align}

\noindent \textit{Step 2:}
\\	
We restrict the infimumn in \eqref{eq_pacb_slow_3} to $ \rho:= p_0 $ given in \eqref{eq_specific_distribution}.
As the quantile loss is 1-Lipschitz, one has that
$
R (\theta) - R (\theta^*) 
\leq
\mathbb{E}	\|X_1\| \|\theta - \theta^* \|
.
$
From Lemma \ref{lema_bound_prior_arnak}, we have, for , that
\begin{align}
\int [ R (\theta) - R (\theta^*) ] p_0 ({\rm d} \theta)
&	\leq
\int \mathbb{E}	\|X_1\| \| \theta- \theta^* \| p_0( {\rm d} \theta) \nonumber
\\
&	\leq
C_{\rm x}
\left( \int \| \theta- \theta^* \|^2 p_0( {\rm d} \theta ) \right)^{1/2} \nonumber
\\
& \leq
C_{\rm x}
\sqrt{	4d\varsigma^2 }
\label{eq_bound_1_1}
\end{align}
and
\begin{align}
\mathcal{K}(p_0 \| \pi)
\leq
4 s^* \log \left(\frac{C_1 }{\varsigma s^*}\right)
+
\log(2)
\label{eq_bound_1_2}
.
\end{align}
Plug-in the bounds in \eqref{eq_bound_1_1} and \eqref{eq_bound_1_2} into inequality \eqref{eq_pacb_slow_3} and take $ \lambda = \sqrt{n} $, one gets with probability at least $ 1- 2\varepsilon $ that
\begin{equation*}
\mathbb{E}_{\theta\sim \hat{\rho}_{\lambda}}
[ R (\theta) ] - R^*
\leq
\inf_{\tau \in (0,C_1/2d)} 
\!
 \left\{ 
C_{\rm x} 2 \varsigma \sqrt{	d }
+  \frac{ C^2}{4\sqrt{n}} 
+ 2 \frac{ 4 s^* \log \left(\frac{C_1 }{\varsigma s^*}\right)
	+
	\log(2) + \log\frac{1}{\varepsilon}}{\sqrt{n} } 
\right\} 
,
\end{equation*}
and the choice $ \varsigma = ( C_{\rm x} n\sqrt{d})^{-1} $ leads to
\begin{align*}
\mathbb{E}_{\theta\sim \hat{\rho}_{\lambda}}
[ R (\theta) ] - R^*
& \leq 
\frac{2}{n} +  \frac{ C^2}{4\sqrt{n}} 
+ 
2 \frac{ 4 s^* \log \left(\frac{C_1 C_{\rm x} n\sqrt{d} }{ s^*}\right)
	+
	\log(2) + \log\frac{1}{\varepsilon}}{\sqrt{n} } 
\\ 
& \leq
C_{_{C,C_1,X}}
\frac{ s^* \log \left(\frac{ n\sqrt{d}}{ s^*}\right)  + \log\frac{1}{\varepsilon}
}{\sqrt{n} }
,
\end{align*}
for some constant $ C_{_{C,C_1,X}} > 0 $ depending only on $ C,C_1,C_{\rm x} $.
\\
Now, plug-in the bounds in \eqref{eq_bound_1_1} and \eqref{eq_bound_1_2} into inequality \eqref{eq_bound_in_EXPECT_slow} and take $ \lambda = \sqrt{n} $, one obtain that
\begin{equation*}
\mathbb{E}\, 
\mathbb{E}_{\theta\sim \hat{\rho}_{\lambda}}
[ R (\theta) ] - R^*
\leq
\inf_{\tau \in (0,C_1/2d)} 
\!
\left\{ 
C_{\rm x} 2 \varsigma \sqrt{	d }
+  \frac{ C^2}{ 8 \sqrt{n}} 
+  \frac{ 4 s^* \log \left(\frac{C_1 }{\varsigma s^*}\right)
	+	\log(2) }{\sqrt{n} } 
\right\} 
,
\end{equation*}
and the choice $ \varsigma = ( C_{\rm x} n\sqrt{d})^{-1} $ leads to
\begin{align*}
\mathbb{E}\, 
\mathbb{E}_{\theta\sim \hat{\rho}_{\lambda}}
[ R (\theta) ] - R^*
& \leq 
\frac{2}{n} +  \frac{ C^2}{8\sqrt{n}} 
+ 
 \frac{ 4 s^* \log \left(\frac{C_1 C_{\rm x} n\sqrt{d} }{ s^*}\right)
	+
	\log(2) }{\sqrt{n} } 
\\ 
& \leq
C_{_{C,C_1,X}}
\frac{ s^* \log \left(\frac{ n\sqrt{d}}{ s^*}\right)  
}{\sqrt{n} }
,
\end{align*}
for some constant $ C_{_{C,C_1,X}} > 0 $ depending only on $ C,C_1,C_{\rm x} $.
The proof is completed.
\end{proof}

\begin{proof}[\bf Proof of Proposition \ref{propo_slow}]
	From \eqref{eq_bound_for_randomized_slow}, for any $ \epsilon \in (0,1) $, we have that	
	\begin{align*}
	\mathbb{E} \Biggl[ \int \exp \left\{ 
	\lambda [R( \theta )-R^* ] 
	-\lambda[r_n( \theta )-r_n^* ]
	- 
	\log \left[\frac{d\hat{\rho}_{\lambda}}{d \pi} (\theta)  \right]
	-
	\frac{\lambda^2 C^2 }{8n} 
	- 
	\log\frac{1}{\epsilon}
	\right\}
	\hat{\rho}_{\lambda}(d \theta)
	\Biggr]
	\leq 
	\epsilon
	.
	\end{align*}
	Thus, using the elementary inequality $\exp(x) \geq \mathbf{1}_{\mathbb{R}_{+}}(x) $ we obtain, with probability at most
	$\epsilon $,
	\begin{align*}
	\lambda [R( \theta )-R^* ] 
	\geq
	\lambda[r_n( \theta )-r_n^* ]
	+
	\log \left[\frac{d\hat{\rho}_{\lambda}}{d \pi} (\theta)  \right]
	+
	\frac{\lambda^2 C^2 }{8n} 
	+
	\log\frac{1}{\epsilon}
	,
	\end{align*}
	where the probability is evaluated with respect to the distribution $\mathbf
	P^{\otimes n}$ of the
	data {\it and} the conditional probability measure $\hat \rho_{\lambda} $.	Taking the complement, with probability at least $ 1-\epsilon $, one has for $ \lambda>0 $ that
	\begin{align*}
	R( \theta )-R^* 
	\leq
	r_n( \theta )-r_n^* 
	+
	\frac{\log \left[\frac{d\hat{\rho}_{\lambda}}{d \pi} (\theta)  \right]
		+
		\frac{\lambda^2 C^2 }{8n} 
		+
		\log\frac{1}{\epsilon}
	}{\lambda}
	.
	\end{align*}
	Note that
	\begin{align*}
	\log\left(
	\frac{\mbox{d}\hat{\rho}_{\lambda}}{\mbox{d}\pi}(\theta)
	\right)
	= 
	\log
	\frac{\exp \left [-\lambda
		r_{n}(\theta)\right]
	}{ \int \exp\left[-\lambda r_{n}(\theta)\right]
		\mbox{d}\pi(\theta) }
	= 
	-\lambda r_{n}(\theta) - \log \int
	\exp\left[-\lambda r_{n}(\theta)\right] \mbox{d}\pi(\theta)
	,
	\end{align*}
	thus, we obtain with probability at least $ 1-\epsilon $ that
	\begin{align*}
	R( \theta )-R^* 
	\leq
	- \frac{1}{\lambda} \log \int
	\exp\left[-\lambda r_{n}(\theta)\right] \mbox{d}\pi(\theta) - r_n^* 
	+
	\frac{	\frac{\lambda^2 C^2 }{8n} 
		+
		\log\frac{1}{\epsilon}
	}{\lambda}
	.
	\end{align*}
	Now, using Lemma \ref{lemma:dv}, we get with probability at least $ 1-\epsilon $ that
	\begin{align*}
	R( \theta )-R^* 
	\leq
	\inf_{\rho\in\mathcal{P}(\Theta)}  
	\left\{ 
	\mathbb{E}_{\theta\sim\rho}[ r(\theta) -  r(\theta^*)] +  \frac{\lambda C^2}{8n} + \frac{	\mathcal{K}(\rho\|\pi) + \log\frac{1}{\varepsilon}}{\lambda} 
	\right\} 
	.
	\end{align*}	
	Combining the above inequality wit \eqref{eq_pacb_slow_2}, we get with probability at least $ 1- 2\epsilon $ that
	\begin{align}
	\label{eq_randomized_slow}
	R( \theta )-R^* 
	\leq
	\inf_{\rho\in\mathcal{P}(\Theta)}  \left\{ 
	\mathbb{E}_{\theta\sim\rho}
	[ R(\theta) - R(\theta^*)] 
	+  2\frac{\lambda C^2}{8n} 
	+ 2 \frac{	\mathcal{K} (\rho\|\pi) + \log\frac{1}{\varepsilon}}{\lambda} 
	\right\} 
	.
	\end{align}	
	We restrict the infimumn in \eqref{eq_randomized_slow} to $ \rho:= p_0 $ given in \eqref{eq_specific_distribution}.
	Using Lemma \ref{lema_bound_prior_arnak}, we can plug-in the bounds in \eqref{eq_bound_1_1} and \eqref{eq_bound_1_2} into inequality \eqref{eq_randomized_slow}, with $ \lambda = \sqrt{n} $, one gets that	
	\begin{equation*}
	R (\theta)  - R^*
	\leq
	\inf_{\tau \in (0,C_1/2d)} 
	\!
	\left\{ 
	C_{\rm x} 2 \varsigma \sqrt{	d }
	+  \frac{ C^2}{4\sqrt{n}} 
	+ 2 \frac{ 4 s^* \log \left(\frac{C_1 }{\varsigma s^*}\right)
		+
		\log(2) + \log\frac{1}{\varepsilon}}{\sqrt{n} } 
	\right\} 
	,
	\end{equation*}
	and the choice $ \varsigma = ( C_{\rm x} n\sqrt{d})^{-1} $ leads to
	\begin{align*}
	R (\theta) - R^*
	& \leq 
	\frac{2}{n} +  \frac{ C^2}{4\sqrt{n}} 
	+ 
	2 \frac{ 4 s^* \log \left(\frac{C_1 C_{\rm x} n\sqrt{d} }{ s^*}\right)
		+
		\log(2) + \log\frac{1}{\varepsilon}}{\sqrt{n} } 
	\\ 
	& \leq
	C_{_{C,C_1,X}}
	\frac{ s^* \log \left(\frac{ n\sqrt{d}}{ s^*}\right)  + \log\frac{1}{\varepsilon}
	}{\sqrt{n} }
	,
	\end{align*}
	for some constant $ C_{_{C,C_1,X}} > 0 $ depending only on $ C,C_1,C_{\rm x} $.
\end{proof}

\subsection{Proof for fast rate}

\begin{proof}[\bf Proof of Theorem \ref{thm_main1}]

As the quantile loss is 1-Lipschitz, one has that
\begin{align*}
\mathbb{E} \{ (\ell_{\tau,i}(\theta) - \ell_{\tau,i}(\theta^*) )^2 \}
\leq
\mathbb{E}	|X_i^\top( \theta - \theta^*) |^2
.
\end{align*}
where $ \ell_{\tau,i} (\theta) :=  \ell_\tau (Y_i , X_i^\top \theta)  $. Thus, from Assumption \ref{assum_bernstein}, one obtains that
\begin{align*}
\mathbb{E} \{ (\ell_{\tau,i}(\theta) - \ell_{\tau,i}(\theta^*) )^2 \}
\leq
	K [ R (\theta) - R^* ]
.
\end{align*}
This means that the required assumption of Theorem \ref{thm:oracle:bound} is satisfied and we can apply it.

	We define the following distribution as a translation of the prior $ \pi $,
	\begin{equation}
	\label{eq_specific_distribution}
	p_0(\theta) 
	\propto 
	\pi (\theta - \theta^*)\mathbf{1}_{
	\{ \| \theta - \theta^*\|_1 \leq 2d\varsigma \} 
}
	.
	\end{equation}
	We restrict the infimumn in \eqref{eq_oracle_in_berns} to $ \rho:= p_0 $ given in \eqref{eq_specific_distribution}.
Using Lemma \ref{lema_bound_prior_arnak}, we can plug-in the bounds in \eqref{eq_bound_1_1} and \eqref{eq_bound_1_2} into inequality \eqref{eq_oracle_in_berns}, one gets that
	\begin{equation*}
\mathbb{E} [
\mathbb{E}_{\theta\sim \hat{\rho}_{\lambda}}
[ R (\theta) ] ] - R^*
\leq
2
\inf_{\tau \in (0,C_1/2d)} 
\!
\left\{  
	C_{\rm x} 2\varsigma\sqrt{ d }
+ 
C_{_{K,C}} \frac{ 4 s^* \log \left(\frac{C_1 }{\varsigma s^*}\right)
	+
	\log(2)
}{n} \right\}
,
\end{equation*}
	and the choice $ \varsigma = ( C_{\rm x} n\sqrt{d})^{-1} $ leads to
\begin{align*}
\mathbb{E} [
\mathbb{E}_{\theta\sim \hat{\rho}_{\lambda}}
[ R (\theta) ] ] - R^*
& \leq
2
\left\{  
\frac{4}{n} 
+ 
C_{_{K,C}} \frac{2 s^* \log \left(\frac{ C_{\rm x} C_1 n\sqrt{d}}{ s^*}\right)
	+
	\log(2)
}{n} \right\}
\\ 
& \leq
C_{_{K,C,X}}
\frac{ s^* \log \left(\frac{ n\sqrt{d}}{ s^*}\right)
}{n}
,
\end{align*}
for some constant $ C_{_{K,C,X}} > 0 $ depending only on $ K,C,C_{\rm x} $. 

\underline{To obtain the bound in probability:}
\\
Fix $\theta\in\Theta$, for $ \ell_{i}(\theta)= \ell_\tau (Y_i , X_i^\top \theta)  $, and apply Bernstein Lemma \ref{lemma:bernstein} with $U_i= \ell_i(\theta^*) - \ell_{i}(\theta)  $. We obtain for any $t>0$ that
$
\mathbb{E}  {\rm e}^{ t n 
	[R(\theta) - R(\theta^*) -r(\theta) + r(\theta^*)] }  
\leq 
{\rm e}^{ g\left(C t\right) n t^2  {\rm Var} (U_i)}.
$
We put $t=\lambda/n$,
and note that
\begin{align*}
{\rm Var} (U_i)
 \leq 
 \mathbb{E} (U_i^2)
 =
 \mathbb{E} \left\{[\ell_i(\theta^*) - \ell_{i}(\theta)]^2 \right\}
 \leq
 \mathbb{E}	|X_i^\top( \theta - \theta^*) |^2
 \leq 
 K \left[ R(\theta) - R(\theta^*) \right]
\end{align*}
thanks to Assumption \ref{assum_bernstein}. Thus we get:
\begin{equation*}
\mathbb{E}  {\rm e}^{ \lambda 	[R(\theta) - R(\theta^*) -r(\theta) + r(\theta^*)]  } 
\leq 
{\rm e}^{ g\left(\frac{\lambda C}{n}\right) \frac{\lambda^2}{n} K \left[ R(\theta) - R(\theta^*) \right] }
.
\end{equation*}
Rearrange the terms and subsequently integrate the resulting expression with respect to \(\pi\) and then, apply Fubini's theorem to obtain the following result:
\begin{equation*}
\mathbb{E}  \mathbb{E}_{\theta\sim\pi} 
\left( {\rm e}^{\lambda \left\{ \left[ 1- K g\left(\frac{\lambda C}{n}\right) \frac{\lambda}{n} \right]\left[ R(\theta) - R(\theta^*)\right] -r(\theta) + r(\theta^*)\right\} } \right)
\leq 1.
\end{equation*}
and we apply Lemma~\ref{lemma:dv} and multiply both sides by $ \varepsilon >0 $ to get:
\begin{equation*}
\mathbb{E} \left[ {\rm e}^{ 
	\lambda 
	\sup_{\rho\in\mathcal{P}(\Theta)} 
	\left\{
		 \mathbb{E}_{\theta\sim\rho}
	\left\{
	\left[ 1- K g\left(\frac{\lambda C}{n}\right) \frac{\lambda}{n} \right] [R(\theta) - R(\theta^*)] -r(\theta) + r(\theta^*) \right\}
	-	
	\mathcal{K}(\rho\|\pi) - \log(1/\varepsilon)
	\right\} 
 } \right] 
\leq 
\varepsilon
.
\end{equation*} 
Uses Chernoff bound  to get
\begin{equation*}
\mathbb{P} \left[ \sup_{\rho\in\mathcal{P}(\Theta)} \lambda \mathbb{E}_{\theta\sim\rho}
\left\{
\left[ 1- K g\left(\frac{\lambda C}{n}\right) \frac{\lambda}{n} \right] 
[R(\theta) - R(\theta^*)] -r(\theta) + r(\theta^*) 
\right\}
-	\mathcal{K}(\rho\|\pi)  > \log\frac{1}{\varepsilon} \right] 
\leq 
\varepsilon.
\end{equation*} 
Take the complement and rearranging terms give, with probability at least $ 1- \varepsilon $
\begin{equation*}
\mathbb{E}_{\theta\sim\rho} 
\left[ 1- K g\left(\frac{\lambda C}{n}\right) \frac{\lambda}{n} \right] 
[R(\theta) - R(\theta^*)] 
\leq
\mathbb{E}_{\theta\sim\rho}[ r(\theta) -  r(\theta^*)] +  \frac{	\mathcal{K}(\rho\|\pi) + \log\frac{1}{\varepsilon}}{\lambda} 
.
\end{equation*}
using Lemma \ref{lemma:dv} again, we get that with probability at least $ 1- \varepsilon $
\begin{multline}
\label{eq_pacb_fast_1}
\mathbb{E}_{\theta \sim \hat{\rho}_\lambda}
\left[ 1- K g\left(\frac{\lambda C}{n}\right) \frac{\lambda}{n} \right] 
[ R(\theta) - R(\theta^*)] 
\\
\leq
\inf_{\rho\in\mathcal{P}(\Theta)}  \left\{ 
\mathbb{E}_{\theta\sim\rho}[ r(\theta) -  r(\theta^*)] + 
 \frac{	\mathcal{K}(\rho\|\pi) + \log\frac{1}{\varepsilon}}{\lambda} 
\right\} 
.
\end{multline}	
Now, with exactly the same argument but with an application to $ -U_i $ rather than $ U_i $, one obtains that  with probability at least $ 1- \varepsilon $
\begin{equation}
\label{eq_pacb_fast_2}
\mathbb{E}_{\theta\sim\rho}[ r(\theta) -  r(\theta^*)]
\leq
\mathbb{E}_{\theta\sim\rho}
\left[ 1+ K g\left(\frac{\lambda C}{n}\right) \frac{\lambda}{n} \right] 
[ R(\theta) - R(\theta^*)] 
+ 
\frac{	\mathcal{K}(\rho\|\pi) + \log\frac{1}{\varepsilon}}{\lambda} 
.
\end{equation}
Combining \eqref{eq_pacb_fast_1} and \eqref{eq_pacb_fast_2}, we get that
with probability at least $ 1- 2\varepsilon $
\begin{multline*}
\mathbb{E}_{\theta \sim \hat{\rho}_\lambda}
\left[ 1- K g\left(\frac{\lambda C}{n}\right) \frac{\lambda}{n} \right] 
[ R(\theta) - R(\theta^*)] 
\\
\leq
\inf_{\rho\in\mathcal{P}(\Theta)}  \left\{ 
\mathbb{E}_{\theta\sim\rho}
\left[ 1+ K g\left(\frac{\lambda C}{n}\right) \frac{\lambda}{n} \right] 
[ R(\theta) - R(\theta^*)] 
+ 
2 \frac{	\mathcal{K} (\rho\|\pi) + \log\frac{1}{\varepsilon}}{\lambda} 
\right\} 
.
\end{multline*} 
From this point, we assume \(\lambda\) is such that \(\left[ 1 - K g\left(\frac{\lambda C}{n}\right) \frac{\lambda}{n} \right] > 0\). As a result,
\begin{multline*}
\mathbb{E}_{\theta \sim \hat{\rho}_\lambda}
[ R(\theta) - R(\theta^*)] 
\\
\leq
\inf_{\rho\in\mathcal{P}(\Theta)}  \left\{ 
\mathbb{E}_{\theta\sim\rho}
\frac{ \left[ 1+ K g\left(\frac{\lambda C}{n}\right) \frac{\lambda}{n} \right]  }{ \left[ 1- K g\left(\frac{\lambda C}{n}\right) \frac{\lambda}{n} \right]  }
[ R(\theta) - R(\theta^*)] 
+ 
2 \frac{ \mathcal{K} (\rho\|\pi) + \log\frac{1}{\varepsilon}}{ \lambda \left[ 1- K g\left(\frac{\lambda C}{n}\right) \frac{\lambda}{n} \right]  } 
\right\} 
.
\end{multline*}  
In particular, take $\lambda = n/\max(2K,C) $. One has that: $\lambda \leq n/(2K) \Rightarrow K \lambda / n \leq 1/2$ and $\lambda \leq n/C \Rightarrow g(\lambda C/n) \leq g(1) \leq 1 $, so
$$  
1 - K g\left(\frac{\lambda C}{n}\right) \frac{\lambda}{n} 
\geq \frac{1}{2} ; \quad
1 + K g\left(\frac{\lambda C}{n}\right) \frac{\lambda}{n} 
\leq \frac{3}{2} 
.
$$
Thus, with probability at least $ 1- 2\varepsilon $, 
\begin{align}
\label{eq_pacb_fast_3}
\mathbb{E}_{\theta \sim \hat{\rho}_\lambda}
[ R(\theta) - R(\theta^*)] 
\leq
\inf_{\rho\in\mathcal{P}(\Theta)}  \left\{ 
\mathbb{E}_{\theta\sim\rho}
3
[ R(\theta) - R(\theta^*)] 
+ 
4  \frac{ \mathcal{K} (\rho\|\pi) + \log\frac{1}{\varepsilon} }{ n / \max(2K,C) } 
\right\} 
.
\end{align} 
	We restrict the infimumn in \eqref{eq_pacb_fast_3} to $ \rho:= p_0 $ given in \eqref{eq_specific_distribution}.
Using Lemma \ref{lema_bound_prior_arnak}, we can plug-in the bounds in \eqref{eq_bound_1_1} and \eqref{eq_bound_1_2} into inequality \eqref{eq_pacb_fast_3}, one gets that
\begin{equation*}
	\mathbb{E}_{\theta\sim \hat{\rho}_{\lambda}}
	[ R (\theta) ]  - R^*
	\leq
	\inf_{\tau \in (0,C_1/2d)} 
	\!
	\left\{  
	6 C_{\rm x} \varsigma\sqrt{ d }
	+ 
	4 C_{_{K,C}} \frac{ 4 s^* \log \left(\frac{C_1 }{\varsigma s^*}\right)
		+
		\log(2) + \log\frac{1}{\varepsilon}
	}{n} \right\}
	,
\end{equation*}
	and the choice $ \varsigma = ( C_{\rm x} n\sqrt{d})^{-1} $ leads to
\begin{align*}
	\mathbb{E}_{\theta\sim \hat{\rho}_{\lambda}}
	[ R (\theta) ]  - R^*
&	\leq
\frac{6}{n}
	+ 
	4 C_{_{K,C}} \frac{ 4 s^* \log \left(\frac{C_1 C_{\rm x} n \sqrt{d} }{ s^*}\right)
		+
		\log(2) + \log\frac{1}{\varepsilon}
	}{n} 
\\ 
& \leq
C_{_{K,C,X}}
\frac{ s^* \log \left(\frac{ n\sqrt{d}}{ s^*}\right) + \log (1/\varepsilon) 
}{n}
,
\end{align*}
for some constant $ C_{_{K,C,X}} > 0 $ depending only on $ K,C,C_{\rm x} $. 	
The proof is completed.
\end{proof}

\begin{proof}[\bf Proof of Proposition \ref{propo_fastrate}]

From Assumption \ref{assum_bernstein}, we have that
	\begin{equation*}
\mathbb{E} \{ 
|  X^\top (\theta - \theta^*) |^2 \}
\leq 
\mathbb{E} K [ R (\theta) - R(\theta^*) ]
.
\end{equation*}
Integrating with respect ot $ \hat{\rho}_{\lambda} $ and using Fubini's theorem, we gets that 
	\begin{equation*}
\mathbb{E} 	\mathbb{E}_{\theta\sim \hat{\rho}_{\lambda}} \{ 
|  X^\top (\theta - \theta^*) |^2 \}
\leq 
\mathbb{E} 	\mathbb{E}_{\theta\sim \hat{\rho}_{\lambda}} K [ R (\theta) - R(\theta^*) ]
,
\end{equation*}
using now the results from Theorem \ref{thm_main1}, we obtain that
	\begin{equation*}
\mathbb{E} 	\mathbb{E}_{\theta\sim \hat{\rho}_{\lambda}} \{ 
|  X^\top (\theta - \theta^*) |^2 \}
\leq
C_{_{K,C,X}}
\frac{ s^* \log \left(\frac{ n\sqrt{d}}{ s^*}\right) 
}{n}
,
\end{equation*}
for some constant $ C_{_{K,C,X}} > 0 $ depending only on $ K,C,C_{\rm x} $. 	

With a similar argument but using now Assumption \ref{assum_eigen}, 
	\begin{equation*}
	\mathbb{E} 
\mathbb{E}_{\theta\sim \hat{\rho}_{\lambda}} \{ 
|  \theta - \theta^* |^2 \}
\leq
C_{_{\kappa,K,C,X}}
\frac{ s^* \log \left(\frac{ n\sqrt{d}}{ s^*}\right)
}{n}
,
\end{equation*}
for some constant $ C_{_{\kappa,K,C,X}} > 0 $ depending only on $ \kappa, K,C,C_{\rm x} $. 	
\end{proof}

\subsection{Lemmas}

First we state a general PAC-Bayesian relative bound in expectation. See Theorem 4.3 in \cite{alquier2021user}.
\begin{theorem}[Theorem 4.3 in \cite{alquier2021user}]
	\label{thm:oracle:bound}
	Assume Assumption \ref{assum_bernstein} is satisfied. Take $\lambda= n/C_{_{K,C}}, C_{_{K,C}} : = \max(2K,C) $, we have:
	\begin{equation}
	\label{eq_oracle_in_berns}
	\mathbb{E} \mathbb{E}_{\theta\sim\hat{\rho}_{\lambda}} [R(\theta) ]- R^*
	\leq 
	2  \inf_{\rho\in\mathcal{P}(\Theta)}  \left\{  \mathbb{E}_{\theta\sim\rho} [R(\theta) ]- R^* 
	+ 
	C_{_{K,C}}	\frac{ \mathcal{K}(\rho\| \pi) }{n} \right\}.
	\end{equation}
\end{theorem}

\begin{lemma}[Hoeffding's inequality]
	\label{lemma:hoeffding}
	Let $U_1,\dots,U_n$ be independent random variables taking values in an interval $[a,b]$. Then, for any $t>0$,
	$$ \mathbb{E} 
	{\rm e}^{t \sum_{i=1}^n [ U_i - \mathbb{E}(U_i)]}  \leq {\rm e}^{\frac{n t^2 (b-a)^2}{8}}. $$
\end{lemma}
\begin{proof}
	The proof can be found for example in Chapter 2 of \cite{MR2319879}.
\end{proof}

\begin{lemma}[Bernstein's inequality]
	\label{lemma:bernstein}
	Let $g$ denote the Bernstein function defined by $g(0) = 1$ and, for $x\neq 0$,
	$$ g(x) = \frac{{\rm e}^x - 1 - x}{x^2}. $$
	Let $U_1,\dots,U_n$ be i.i.d random variables such that $\mathbb{E}(U_i)$ is well defined and $U_i- \mathbb{E}(U_i) \leq C$ almost surely for some $C\in\mathbb{R}$. Then
	$$
	\mathbb{E}\left( {\rm e}^{t \sum_{i=1}^n [U_i - \mathbb{E}(U_i) ] } \right)
	\leq {\rm e}^{ g\left(C t\right) n t^2 {\rm Var}(U_i) }.
	$$
\end{lemma}
\begin{proof}
	For a proof, see Theorem 5.2.1 in~\cite{catoni2004statistical}.
\end{proof}

\begin{lemma}[Donsker and Varadhan's variational formula, \cite{catonibook}]
	\label{lemma:dv}
	For any measurable, bounded function $h:\Theta\rightarrow\mathbb{R}$ we have:
	\begin{equation*}
	\log \mathbb{E}_{\theta\sim\pi}\left[{\rm e}^{h(\theta)} \right] =\sup_{\rho\in\mathcal{P}(\Theta)}\Bigl[\mathbb{E}_{\theta\sim\rho}[h(\theta)] -	\mathcal{K} (\rho\|\pi)\Bigr].
	\end{equation*}
	Moreover, the supremum with respect to $\rho$ in the right-hand side is
	reached for the Gibbs measure
	$\pi_{h}$ defined by its density with respect to $\pi$
	\begin{equation*}
	\frac{{\rm d}\pi_{h}}{{\rm d}\pi}(\theta) =  \frac{{\rm e}^{h(\theta)}}
	{ \mathbb{E}_{\vartheta\sim\pi}\left[{\rm e}^{h(\vartheta)} \right] }.
	\end{equation*}
\end{lemma}

\begin{lemma}
	\label{lema_bound_prior_arnak}
	Let $p_0 $ be the probability measure defined by (\ref{eq_specific_distribution}). If
	$d\geq 2$ then
	$$
	\int_\Omega \| \theta- \theta^* \|^2 p_0({\rm d} \theta)
	\leq
	4d\varsigma^2 
	,
	$$
	and
	$$
	\mathcal{K}(p_0 \| \pi)
	\leq
	4 s^* \log \left(\frac{C_1 }{\varsigma s^*}\right)
	+
	\log(2)
	.
	$$
\end{lemma}
\begin{proof}
	The proof can be found in \cite{mai2023high}, which utilizes results from \cite{dalalyan2012mirror}.
\end{proof}

\end{document}